# Lower Bound Bayesian Networks – An Efficient Inference of Lower Bounds on Probability Distributions in Bayesian Networks

**Daniel Andrade**
Department of Computer Science
University of Tokyo
7-3-1, Hongo, Bunkyo-ku, Tokyo, Japan
daniel.andrade@is.s.u-tokyo.ac.jp

**Bernhard Sick**
Computationally Intelligent Systems (CIS)
University of Passau
94030 Passau, Germany
sick@fim.uni-passau.de

## Abstract

We present a new method to propagate lower bounds on conditional probability distributions in conventional Bayesian networks. Our method guarantees to provide outer approximations of the exact lower bounds. A key advantage is that we can use any available algorithms and tools for Bayesian networks in order to represent and infer lower bounds. This new method yields results that are provable exact for trees with binary variables, and results which are competitive to existing approximations in credal networks for all other network structures. Our method is not limited to a specific kind of network structure. Basically, it is also not restricted to a specific kind of inference, but we restrict our analysis to prognostic inference in this article. The computational complexity is superior to that of other existing approaches.

## 1 INTRODUCTION

A *Bayesian network* is a popular means to represent joint probability distributions over a set of random variables. It consists of a graphical network which specifies dependencies between random variables by a directed graph, and of conditional probability tables (CPT) which specify for each variable a conditional probability distribution. Calculating the marginal probability distribution of a variable, given a set of observed variables (variables with evidence), is called *inference* in a Bayesian network.

A special case of inference is *prognostic inference* which is the calculation of $P(X|E_1, \ldots, E_v)$ if all evidence nodes $E_1, \ldots, E_v$ are (direct or indirect) predecessors of node $X$ and a predecessor of an evidence node is also an evidence node. Calculating any marginal distribution $P(X)$ without evidence is, therefore, considered as a special case of prognostic inference. The inference result can be used, for example, to determine expected profits, or to make important decisions. Prognostic reasoning can be an important tool, for instance, in medicine where temporal dependencies are explicitly modeled with a Bayesian network. For example, the outcome of a treatment, measured by the life expectancy of a patient, can be modeled as a consequence of the observed symptoms before treatment, and a sequence of treatments [Lucas et al., 2004].

Obviously, outcomes and decisions of an inference are sensitive to the choice of the CPT in the Bayesian network. Since most of these probabilities are typically estimated, it makes sense to check how sensitive the outcome is regarding a change of these probabilities. A method which expresses the posterior as a function of a node probability is described in [Kjærulff and van der Gaag, 2000]. Various other types of approaches have been suggested to tackle this problem. A general idea is to model the uncertainty about the probabilities of each node and to propagate this uncertainty through the Bayesian network. A natural extension is to define a continuous probability distribution over the discrete probability distributions at each node (second-order probabilities). In [Borsotto et al., 2006] an analytic solution is derived to propagate expectation and variance of second-order probabilities. However, the computational complexity of this approach is high. Another approach to model uncertainty about a discrete probability distribution is to define probability intervals (or, more general, convex sets) instead of point probabilities which leads to the idea of *credal networks* [de Campos and Cozman, 2005]. How could probability intervals be found? One could, for example, ask several experts to estimate a probability interval. For that task it turns out that the *imprecise Dirichlet model* is a suitable means which has several desirable properties (see [Andrade et al., 2008]). Assume, for example, we want to find a lower and upper bound for the prior $P(X)$. Further assume that $n_i$ statements were given by experts favoring the statement $[X = x_i]$, and the total number of statements is $n$, then the lower and upper bounds for $P(X = x_i)$ can be calculated by

$$\underline{P}(X = x_i) = \frac{n_i}{d + n},$$
$$\overline{P}(X = x_i) = \frac{n_i + d}{d + n},$$

respectively, where $d$ is a hyperparameter which is set to a strictly positive value, preferably 1 or 2 (see [Walley, 1996]). If we define the constant $\frac{d}{d+n}$ as the *degree of ignorance* about the probability distribution $P(X)$, we can see that the upper bound is actually induced by the lower bound. That is, given the lower bound for any $P(X = x_i)$, we can calculate the upper bound by adding the degree of ignorance. The consequence is that in this case it is sufficient to have a framework which allows to specify lower bounds (and not explicit lower and upper bounds). One might think that being able to propagate lower bounds in a Bayesian network without error is sufficient to infer exact upper bounds by calculating the induced upper bounds. Unfortunately, this is not true since an exact upper bound can be lower than an upper bound induced by the correponding exact lower bounds. But it is easy to see that in the case where all nodes have only two states, the exact upper bound always equals the induced upper bound.

Basically, the idea of specifying only lower bounds inspired us to define the new approach outlined in this article, which we call the *Lower Bound Bayesian Network* (*LBBN*) method. LBBN provide an elegant and efficient solution to approximate exact lower bounds which can be run on any Bayesian network structure. The inferred lower bound and the induced upper bound are guaranteed to form an interval which includes the exact lower and upper bound, i.e., we guarantee an *outer approximation*. In the case that the network structure is a tree and all nodes are binary, our approach yields exact solutions for lower and upper bounds—and it does this faster than the 2U algorithm [Ide and Cozman, 2008]. Whereas existing approximations for interval propagation in graphical models often set limitations on the structure of the graphical model or on the number of states, our method can be run on any Bayesian network using any existing algorithm for inference in standard Bayesian networks. Therefore, our method is complementary to some existing methods (e.g., in the case of prognostic inference in multi-connected networks) and competing with some existing methods (e.g., in the case of prognostic inference in binary multi-connected networks). In cases where it is competing, we will demonstrate (Sections 4.3 and 5) that our method always provides the best computational complexity, and it even leads to exact results in trees with binary variables. In all other cases it yields the second-best accuracy among the other existing approximations analyzed in this article.

In the following section we will give a short overview of the existing work which is closely related to our method. In Section 3 we will introduce LBBN in detail. In Section 4 we will sketch the proof that our method delivers outer approximations for lower and upper bounds which is a very desirable property, since only this property allows for making prudent decisions based on the approximated estimates. Furthermore, we will sketch the proof that our method yields exact solutions when the network is a tree and all variables are binary. In Section 4.3 we will show the superiority of LBBN in terms of computational complexity, and in Section 5 we demonstrate with some simulation experiments that our method is competitive to approximations for credal networks in terms of accuracy. Section 6 finally summarizes the major findings.

## 2　RELATED WORK

Currently, the most important means for representing and propagating lower and upper bounds on probabilities in Bayesian networks are *credal networks*. They generalize the concept of Bayesian networks by propagating convex sets of discrete probability distributions instead of point probabilities (see, e.g., [de Campos and Cozman, 2005]). However, inference even in polytrees, except if all nodes have only two states, is shown to be NP-hard [de Campos and Cozman, 2005]. Therefore, various approximations have been suggested (see [da Rocha et al., 2003, Ide and Cozman, 2008, Tessem, 1992]). We will compare the performance of inference methods in credal networks to our method in Sections 4.3 and 5.

Regarding the idea to concentrate on specifying only lower bounds instead of lower and upper bounds, our method is closely related to the work of Fertig and Breese [Fertig and Breese, 1990]. Their method yields outer approximations, too. However, the run-time of their algorithm is exponential with respect to the number of nodes (see Section 4.3). We will also make use of one result of Fertig and Breese's work in Section 4.2 in order to prove that our method yields exact lower bounds in trees with binary variables (and, thus, exact upper bounds, too).

## 3　LOWER BOUND BAYESIAN NETWORKS (LBBN)

In this section, Lower Bound Bayesian Networks are introduced. LBBN provide a novel way to approximate lower bounds. They deliver outer approximations such as the methods which were mentioned in the preceding section. In contrast to those, LBBN does not require to implement new algorithms or to extend given data structures. The reason is that they are actually standard Bayesian networks, where only the interpretation of the entries in the CPT is modified.

### 3.1 DEFINITION OF LBBN

First, let us briefly recall the definition of a Bayesian network. It is defined by a directed non-cyclical graph. Furthermore, for each node $X_i$ there is a discrete conditional probability distribution $P(X_i|\Pi(X_i))$, where $\Pi(X_i)$ are the predecessor nodes of node $X_i$. The conditional probability distribution for each node is defined in the conditional probability tables (CPT). Let us refer to the state space of $X_i$ by $S_i$.

In an LBBN, the structure equals the structure of a Bayesian network, however, the meaning of the entries in the CPT is changed. First, each entry of the CPT $P(X = x|\Pi(X))$ is replaced by a lower bound $\underline{P}(X = x|\Pi(X))$. After this step, the entries in a CPT do not sum up to one anymore. The mass $1 - \sum_x \underline{P}(X = x|\Pi(X))$ is the *degree of ignorance* and it can also be interpreted as a kind of free mass which is not assigned to any state of $X$. In order to recreate probabilities, a new state, called $N$ is introduced, which collects this free mass in each CPT. Finally, having adapted the CPT appropriately, any Bayesian network algorithm can be used to infer any marginal distribution in this newly created Bayesian network.

Let us now describe this transformation in some more detail. First, the LBBN has the same structure as the Bayesian network, but each node gets an additional state $N$. The node $X_i$ of the original Bayesian network will be named $X'_i$ in the new Bayesian network, and the state space of node $X'_i$ will be named $S'_i = S_i \cup \{N\}$. The conditional probability distribution for a node $X'_i$ is then defined as follows: First, assume that $x_i \in S_i$. If none of the states of the nodes in $\Pi(X'_i)$ is $N$, we set $P(X'_i = x_i|\Pi(X'_i))$ to the given lower bound of the probability $P(X_i = x_i|\Pi(X_i))$, i.e.,

$$P(X'_i = x_i|\Pi(X'_i)) := \underline{P}(X_i = x_i|\Pi(X_i)).$$

If there are $e$ nodes $X'_{u_1}, \ldots, X'_{u_e} \in \Pi(X'_i)$, $e \geq 1$, with their states being set to $N$, then

$$P(X'_i = x_i|\ldots, X'_{u_1} = N, X'_{u_2} = N, \ldots, X'_{u_e} = N, \ldots) := \min_{x_{u_1} \in S_{u_1}, \ldots, x_{u_e} \in S_{u_e}} \underline{P}(X_i = x_i|\Pi(X_i)),$$

where $P(X'_i = x_i|\ldots, X'_{u_1} = N, X'_{u_2} = N, \ldots, X'_{u_e} = N, \ldots)$ is the conditional probability associated with node $X'_i$, given that the predecessors $X'_{u_1}, \ldots, X'_{u_e}$ are set to $N$ and all its other predecessors are set to a state which is not $N$. $X'_u = N$ means that node $X_u$ can be in *any* state $S_u$. As a consequence, the previously defined probability can be interpreted as the minimal chance of being in state $X_i = x_i$, given our ignorance about the current states of $X_{u_1}, X_{u_2}, \ldots, X_{u_e}$. This interpretation of probability mass in state $N$ is comparable to that in Dempster-Shafer theory, when mass is allocated to the whole state space $\Omega$. Finally, consider $x_i$ being $N$. In that case, we set $P(X'_i = x_i|\Pi(X'_i))$ to the free mass, i.e.,

$$P(X'_i = N|\Pi(X'_i)) := 1 - \sum_{x_i \in S_i} P(X'_i = x_i|\Pi(X'_i)).$$

In other words, if we are uncertain about whether we are in a certain state of $X_i$, we represent this by $X_i = N$, meaning we might be in any state of $X_i$. Note that in the case that none of the states in $\Pi(X'_i)$ is $N$, $P(X'_i = N|\Pi(X'_i)) = 1 - \sum_{x_i \in S_i} \underline{P}(X_i = x_i|\Pi(X_i))$.

### 3.2 EXAMPLE

Consider the Bayesian network shown in Figure 1. Each node is binary and the state spaces of the nodes $E$, $F$ and $G$ are $\{e_1, e_2\}$, $\{f_1, f_2\}$, and $\{g_1, g_2\}$, respectively. Given the lower bounds on the conditional probability distributions of each node, we are interested in finding the exact lower bound for $P(G = g_1)$. The lower bounds are defined according to Table 1.

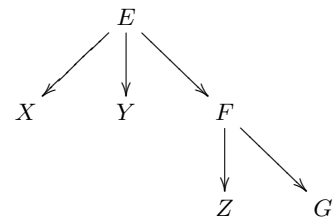

Figure 1: Graph Structure of a Bayesian Network (Example).

Table 1: Lower Bounds on Conditional Probabilities (Example).

| Node $E$ | Node $F$ | Node $G$ |
|---|---|---|
| $\underline{P}(e_1) = 0.6$ | $\underline{P}(f_1|e_1) = 0.4$ | $\underline{P}(g_1|f_1) = 0.7$ |
| $\underline{P}(e_2) = 0.2$ | $\underline{P}(f_2|e_1) = 0.5$ | $\underline{P}(g_2|f_1) = 0.2$ |
| | $\underline{P}(f_1|e_2) = 0.3$ | $\underline{P}(g_1|f_2) = 0.8$ |
| | $\underline{P}(f_2|e_2) = 0.6$ | $\underline{P}(g_2|f_2) = 0.1$ |

First, we create a new Bayesian network which has the same structure, but each node gets an additional state $N$. In order to distinguish between the old nodes and the new nodes with one extra state, we add the symbol $'$ to the nodes, i.e., node $E$, becomes $E'$ and so forth. The CPT of this LBBN are then defined according to Section 3.1 which results in Table 2. Note that probability tables for the nodes $X$, $Y$, and $Z$ are not stated since they are irrelevant for calculating $P(G = g_1)$—these nodes are also called *barren*, see [Shachter, 1990]. We can now determine $P(G' = g_1)$ using any available algorithm for inference in Bayesian networks. In this case we might just sum over all other nodes which results in $P(G' = g_1) = 0.752$. This value is the exact lower bound for the probability $P(G' = g_1)$.

Table 2: Corresponding Entries in the CPT of the LBBN (Example).

| Node $E'$ | Node $F'$ | Node $G'$ |
|---|---|---|
| $P(e_1) = 0.6$ | $P(f_1|e_1) = 0.4$ | $P(g_1|f_1) = 0.7$ |
| $P(e_2) = 0.2$ | $P(f_2|e_1) = 0.5$ | $P(g_2|f_1) = 0.2$ |
| $P(N_E) = 0.2$ | $P(N|e_1) = 0.1$ | $P(N|f_1) = 0.1$ |
|  | $P(f_1|e_2) = 0.3$ | $P(g_1|f_2) = 0.8$ |
|  | $P(f_2|e_2) = 0.6$ | $P(g_2|f_2) = 0.1$ |
|  | $P(N_F|e_2) = 0.1$ | $P(N|f_2) = 0.1$ |
|  | $P(f_1|N_E) = 0.3$ | $P(g_1|N_F) = 0.7$ |
|  | $P(f_2|N_E) = 0.5$ | $P(g_2|N_F) = 0.1$ |
|  | $P(N_F|N_E) = 0.2$ | $P(N_G|N_F) = 0.2$ |

In the following we will show that in certain special cases this procedure always yields exact lower bounds, and in other cases it yields conservative approximations.

## 4 MAJOR PROPERTIES OF LBBN

In this section we prove two important properties of LBBN. The first is that inference in LBBN yields outer approximations, which is a fundamental justification for our method. The second is that we can even deliver exact solutions if we restrict the network to trees with binary variables. We also comment on the computational complexity of inference in LBBN.

### 4.1 OUTER APPROXIMATION

What we want to show is that when calculating $P(X'|E'_1, \ldots, E'_v)$ with any existing Bayesian network inference algorithm, $P(X'|E'_1, \ldots, E'_v)$ is a conservative approximation of the exact lower bound $\underline{P}(X|E_1, \ldots, E_v)$. Note that if $P(X'|E'_1, \ldots, E'_v)$ is a conservative approximation of the exact lower bound, we can easily find a conservative approximation of the exact upper bound $\overline{P}(X|E_1, \ldots, E_v)$ by using the upper bound which is induced by the approximated lower bounds, i.e., $\overline{P}(X = a|E_1, \ldots, E_v) \leq 1 - \sum_{x \in S \setminus \{a\}} P(X' = x|E'_1, \ldots, E'_v)$.

In order to facilitate the notation, let us make the following assumptions without loss of generality. Let the $n$ nodes in the network $X_1, \ldots, X_n$ be ordered such that observed variables $(E_1, \ldots E_v)$ correspond to $X_{r+2}, \ldots, X_n$, and $X$ corresponds to $X_{r+1}$. We assume further that the indices of the other nodes are chosen such that $X_1, \ldots, X_r$ are sorted using their top sort rank. Top sort can be run on the network since a Bayesian network does not contain cycles. For the observed variables we shortly write $\mathbf{E}$ (and when we refer to the corresponding nodes in the LBBN we write $\mathbf{E}'$). Therefore,

$$P(X|\mathbf{E}) = \frac{\sum_{x_1} \cdots \sum_{x_r} \prod_{i=1}^{n} P(X_i|\Pi(X_i))}{\sum_{x_1} \cdots \sum_{x_{r+1}} \prod_{i=1}^{n} P(X_i|\Pi(X_i))} \ .$$

In the general inference case, we could not simplify further, and using our proposed method we would have to find a lower bound for the nominator and an upper bound for the denominator independently. In our future work we must, therefore, investigate the additional approximation error introduced in the case of general inference. Here, we will limit the discussion to prognostic inference, which equals to $X_i \in \mathbf{E} \Rightarrow \Pi(X_i) \in \mathbf{E}$, and we simplify the above equation to:

$$P(X|\mathbf{E}) =$$

$$\frac{\sum_{x_1} \cdots \sum_{x_r} \prod_{i=1}^{n} P(X_i|\Pi(X_i))}{\prod_{i=r+2}^{n} P(X_i|\Pi(X_i)) \sum_{x_1} \cdots \sum_{x_{r+1}} \prod_{i=1}^{r+1} P(X_i|\Pi(X_i))}$$

and thus

$$P(X|\mathbf{E}) = \sum_{x_1} \cdots \sum_{x_r} \prod_{i=1}^{r+1} P(X_i|\Pi(X_i)) \ .$$

Next, we will use that we sorted the indices according to top sort, and factor out the conditional probabilities which leads to

$$P(X|\mathbf{E}) = \sum_{x_1} P(X_1|\Pi(X_1)) \ldots$$
$$\ldots \sum_{x_r} P(X_r|\Pi(X_r)) \cdot P(X_{r+1}|\Pi(X_{r+1})).$$

We can show by induction over $r$ (see Appendix A) that

$$\sum_{x_1 \in S_1} P(X_1|\Pi(X_1)) \ldots$$
$$\ldots \sum_{x_r \in S_r} P(X_r|\Pi(X_r)) \cdot P(X_{r+1}|\Pi(X_{r+1})) \geq$$
$$\sum_{x_1 \in S'_1} P(X'_1|\Pi(X'_1)) \ldots \quad (1)$$
$$\ldots \sum_{x_r \in S'_r} P(X'_r|\Pi(X'_r)) \cdot P(X'_{r+1}|\Pi(X'_{r+1})) \ .$$

Therefore, $P(X|\mathbf{E}) \geq P(X'|\mathbf{E}')$. Finally, $P(X|\mathbf{E}) \geq \underline{P}(X|\mathbf{E}) \geq P(X'|\mathbf{E}')$ which proves our claim.

## 4.2 EXACT LOWER BOUNDS

Making the same assumptions as above and assuming additionally that the given network is a tree and all variables are binary, we will now show that our method yields exact lower bounds, i.e., $P(X'|\mathbf{E}')$ inferred in the LBBN equals $\underline{P}(X|\mathbf{E})$. Note that in this case, we can then easily calculate exact upper bounds using $\overline{P}(X = a|\mathbf{E}) = 1 - \sum_{x \in S \setminus \{a\}} \underline{P}(X = x|\mathbf{E})$ and $|S| = 2$.

Let us use the same notation as above. Since we have a tree and consider only prognostic inference, the only part of the network which is relevant for calculating $P(X|\mathbf{E})$ is the chain from $X_1$ to $X_{r+1}$, where all predecessors of $X_1$ are observed variables (variables with evidence). One way to retrieve the exact lower bound for $P(X_{r+1}|\mathbf{E})$ is to use a sequence of "node reduction" operations [Shachter, 1986, Fertig and Breese, 1990]. A node reduction operation is used to remove a node $Y$ from the network in the following way: Let us assume that the node $Y$ is connected with a directed edge to node $X$ and that $X$ has no other predecessors (which is true since we consider only trees). After removing $Y$, one has to connect its predecessors to its successor $X$ and can calculate the exact lower bound for a new conditional probability associated with $X$ in the following way: Let $\Pi^{new}(X)$ be the new set of predecessors of $X$, i.e., $\Pi^{new}(X) = (\Pi(X) \setminus \{Y\}) \cup \Pi(Y) = \Pi(Y)$. Then,

$$\underline{P}(X|\Pi^{new}(X)) := \\ \underline{P}(X|Y = y^*) \cdot u(Y = y^*|\Pi(Y)) \\ + \sum_{y \in (S_Y \setminus \{y_*\})} \underline{P}(Y = y|\Pi(Y))\underline{P}(X|Y = y) \,,$$

where $y^* := \operatorname{argmin}_{y \in S_Y} \underline{P}(X|Y = y)$ and $u(Y = y^*|\Pi(Y)) := 1 - \sum_{y \neq y_*} \underline{P}(Y = y|\Pi(Y))$. In other words, for a node removal operation it is guaranteed that the resulting lower bound is again the exact lower bound. Since all nodes are assumed to be binary, the exact upper bound must be the induced upper bound. As a consequence, another node removal operation will again yield the exact lower bound, and so forth. Therefore, one way to calculate $\underline{P}(X_{r+1}|\mathbf{E})$ is to perform a sequence of node removal operations starting with $X_1$ and ending with the removal of $X_r$.

If we can show that one node removal operation equals a summation over the corresponding node in the LBBN, it is then easy to show (proof by induction) that a summation over the nodes $X'_1$, ..., $X'_r$ yields the exact lower bound for $P(X|\mathbf{E})$. In the following, for the sake of brevity, we limit the proof to one node removal operation, and assume that $r = 1$. Let us denote with $\underline{P}(X_2|\Pi^{new}(X_2))$ the exact lower bound for $P(X_2|\Pi^{new}(X_2))$, where $\Pi^{new}(X_2)$ denotes the predecessors of $X_2$ after the node removal of $X_1$. In other words, by removing $X_1$ we retrieve $\underline{P}(X_2|\Pi^{new}(X_2))$. Then, $\underline{P}(X_2|\Pi^{new}(X_2)) = \underline{P}(X_2|\mathbf{E})$. Now let us prove that $P(X'_2|\mathbf{E}') = \underline{P}(X_2|\Pi^{new}(X_2))$:

$$P(X'_2|\mathbf{E}') = \sum_{x_1 \in S'_1} P(X'_1 = x_1|\mathbf{E}')P(X'_2|X'_1 = x_1)$$

$$= \sum_{x_1 \in S_1} P(X'_1 = x_1|\mathbf{E}')P(X'_2|X'_1 = x_1)$$
$$+ P(X'_1 = N|\mathbf{E}') \cdot P(X'_2|X'_1 = N)$$

$$\overset{1)}{=} \sum_{x_1 \in S_1} \underline{P}(X_1 = x_1|\mathbf{E})\underline{P}(X_2|X_1 = x_1)$$
$$+ (1 - \sum_{x_1 \in S_1} \underline{P}(X_1 = x_1|\mathbf{E})) \cdot \underline{P}(X_2|X_1 = x_1^*)$$

$$= \sum_{x_1 \in (S_1 \setminus \{x_1^*\})} \underline{P}(X_1 = x_1|\mathbf{E})\underline{P}(X_2|X_1 = x_1)$$
$$+ \underline{P}(X_1 = x_1^*|\mathbf{E}) \cdot \underline{P}(X_2|X_1 = x_1^*)$$
$$+ (1 - \sum_{x_1 \in (S_1 \setminus \{x_1^*\})} \underline{P}(X_1 = x_1|\mathbf{E})) \cdot \underline{P}(X_2|X_1 = x_1^*)$$
$$- \underline{P}(X_1 = x_1^*|\mathbf{E}) \cdot \underline{P}(X_2|X_1 = x_1^*)$$

$$= \sum_{x_1 \in (S_1 \setminus \{x_1^*\})} \underline{P}(X_1 = x_1|\mathbf{E})\underline{P}(X_2|X_1 = x_1)$$
$$+ u(X_1 = x_1^*|\mathbf{E}) \cdot \underline{P}(X_2|X_1 = x_1^*)$$

$$\overset{2)}{=} \underline{P}(X_2|\Pi^{new}(X_2)) \,.$$

In this proof,

1) $x_1^* := \operatorname{argmin}_{x_1 \in S_{X_1}} \underline{P}(X_2|X_1 = x_1)$, and

2) uses the definition of the node removal operation.

The consequence is that $P(X'|\mathbf{E}') = \underline{P}(X|\mathbf{E})$ which proves our claim.

## 4.3 COMPUTATIONAL COMPLEXITY

Table 3 shows the computational complexity of LBBN and various approximations which are available for credal networks. Since an LBBN is just a Bayesian network, we can run any available inference algorithm for Bayesian networks in order to determine lower and upper bounds. If the network structure is a polytree, we can, for example, use Pearl's algorithm [Pearl, 1988]. This way—assuming $s$ is the number of states of a variable and $p$ is the number of predecessors in the network—we get a computational complexity of $O((s+1)^p)$ per state of a node[1] , since each node in the resulting Bayesian network will have $s+1$ states—one extra state for modeling $N$, the degree of ignorance.

---
[1]Thus, for all states of a node we get $O((s+1)^{p+1})$.

Table 3: Computational Complexity (per Node) of Several Approximations for Credal Networks (Categorized by the Structure of the Network and the Number of States per Node).

| | | | |
|---|---|---|---|
| Binary Poly | **LBBN** | **2U** | |
| | $O(3^p)$ | $O(4^p)$ | |
| Polytree | **LBBN** | **A/R** | **A/R+** |
| | $O((s+1)^p)$ | $O(s^p log(s^p))$ | $O(s^{O(s^p)})$ |
| Binary Multi | **LBBN** | **L2U, IPE** | **2V2U** |
| | $O(3^p)$ | $O(4^p)$ | $O(2^G)$ |
| Multi | **LBBN** | | |
| | $O((s+1)^t)$ | | |

The first row shows the complexity of algorithms which can be run efficiently for polytrees with binary variables. Though the A/R [Tessem, 1992] and A/R+ [da Rocha et al., 2003] can be run in this case as well, we excluded them in this row since it is of little interest to compare them in this case with LBBN or 2U. The 2U algorithm (see [Ide and Cozman, 2008]) is actually not an approximation but yields exact solutions in $O(4^p)$ per node. However, our algorithm is faster and yields also exact results when the polytree is a tree.

Another important point that must be mentioned is that inference with LBBN can currently provide the best computational complexity of all outer approximations in polytrees. The computational complexity of A/R, including the log factor, is due to the descending and ascending sorting during the annihilation and reinforcement operations, respectively. The proof for the computational complexity of A/R+ has been omitted for the sake of brevity.

In multi-connected networks, all approximations are NP-hard. Inference in a multi-connected LBBN is NP-hard due to the fact that inference in a multi-connected Bayesian network is NP-hard. Since fast inference in Bayesian networks can be crucial, there are numerous approaches for approximations of marginal distributions in multi-connected Bayesian networks (see, for example, [Dagum and Luby, 1997]). All of them can be used to approximate marginal distributions in LBBN as well. This way, approximative lower bounds in multi-connected networks can be found in polynomial time. Thus, LBBN can be used to infer approximative lower bounds in large multi-connected Bayesian networks, where all other algorithms might be infeasible. In the corresponding row of Table 3 we consider the costs for one node evaluation, whereas several iterations over several nodes might be necessary till convergence (see [Draper and Hanks, 1994]). This also holds for the other algorithms for multi-connected networks which are mentioned in the following.

There are several approximations available for the special case of inference in multi-connected Bayesian networks where all nodes are binary, e.g., L2U, IPE and 2V2U, see [Ide and Cozman, 2008]. The approximations L2U, IPE, and 2V2U are all based on 2U, which explains, for example, the identical complexity of L2U and IPE which contains the factor $4^p$. 2V2U also makes use of 2U. In general, the computational costs for the summation over the Markov blanket with $G$ nodes, which are proportional to $2^G$, will outweigh the factor $4^p$.

The last row in Table 3 shows the computational complexity of the LBBN method for inference in any multi-connected Bayesian network, whereas $t$ is the treewidth. We found some algorithms which can be used to approximate inference in general multi-connected credal networks (e.g., [Cozman et al., 2004, Ide and Cozman, 2005, Alessandro Antonucci and de Campos, 2008]), but we did not find clear complexity statements and performance evaluations for those. Therefore, it must be investigated in the future whether our method can also outperform these approximations in terms of computational complexity or accuracy.

Finally, we want to remark that we excluded the approach from Fertig and Breese from the table since no efficient algorithms are known even for polytrees and, thus, the best computational complexity which can be stated for their algorithm is $O(s^{O(n)})$. This is due to the fact that the algorithm uses a sequence of "node reduction" and "arc reversal" operations [Shachter, 1986] to infer a posterior distribution. Finding an optimal sequence of these operations is in general intractable, even for polytrees.

One can see from Table 3 that in all cases the computational complexity per node is best for the LBBN method. Although comprehensive run-time experiments are necessary to assure the superiority of the LBBN method in terms of actual run-time, we expect that our method outperforms all other methods since any available, highly-optimized implementation for inference in Bayesian networks can be used to infer probabilities in LBBN.

In the following section we will investigate the accuracy of the LBBN method with some simulation experiments. All approximations that were compared to LBBN in this section with regard to computational complexity will be compared to LBBN regarding accuracy.

## 5 EXPERIMENTS

In order to calculate the deviation from the exact solution we used for all experiments Cozman's credal network implementation (see [Cozman, 2002]).

### 5.1 BENCHMARK NETWORK

In [da Rocha et al., 2003], the accuracy of the A/R and the A/R+ algorithms is investigated using the network displayed in Figure 2.[2] In this network, we infer approximations for the lower bounds with our method and then use the induced upper bounds to compare our results to the ones in [da Rocha et al., 2003].

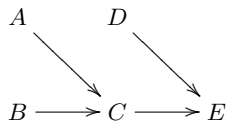

Figure 2: The Inference of $\overline{P}(E = e_1)$ is Used to the Test the Accuracy of A/R and A/R+ in [da Rocha et al., 2003].

Table 4: Results of the Approximation of $\overline{P}(E = e_1)$.

|     | A/R+ | LBBN | A/R  |
| --- | ---- | ---- | ---- |
| MEA | 0.01 | 0.08 | 0.15 |

Each node has 3 states and the credal set for each node is defined by 3 vertices. Choosing random vertices for the definition of each credal set, 15 different instances of this network were created in [da Rocha et al., 2003]. For each of those networks the upper bound for $P(E = e_1)$ was calculated using A/R and A/R+. These approximated upper bounds were then compared to the exact upper bounds. The results can be seen in Table 4. In order to make the testing conditions for the approximation with LBBN close to the ones used in [da Rocha et al., 2003], we randomly generated 3 vertices which were then used to retrieve the lower bounds for the conditional probability distribution of a node. Furthermore, we found that 15 tests are quite few, since the variation is significant when running ten trials each consisting of 15 tests. As a consequence, we decided to run 200 tests in order to get a more stable assessment. The results in Table 4, where MEA denotes mean error in absolute terms, show that LBBN clearly outperforms A/R. The much slower A/R+ (see previous section) is able to produce better results.

---

[2]The actual network in [da Rocha et al., 2003], Figure 1, is larger than the one here. However, the other nodes depicted there are barren for the calculation of the marginal probability $P(E = e_0)$.

### 5.2 RANDOMLY GENERATED NETWORKS

In order to compare our results to the ones presented in [Ide and Cozman, 2008], we randomly generated Bayesian networks using the BNGenerator [Ide, 2004]. Except for the number of nodes, the number of states, and the number of edges, we used the default parameter settings. These parameters were set in order to create five types of multi-connected networks with 10 nodes. The first 3 types have an edge density of 1.2 (i.e., 12 edges), 1.4, and 1.6, and each network contains only binary nodes. The last 2 types have an edge density of 1.2 and 1.4, respectively, and all their nodes have four states. We randomly created 10 different network structures for each type, and for each network structure we created 100 instances with randomly chosen lower bounds. For each instance we calculated the Mean Square Error (MSE). The averages over all instances for all structures of one type are set out in Table 5. We calculated the MSE in accordance with the calculation set out in [Ide and Cozman, 2008]:

$$\sqrt{\frac{1}{Q} \sum_X \sum_{x \in S_X} (\underline{P}^*(x) - \underline{P}(x))^2 + (\overline{P}^*(x) - \overline{P}(x))^2},$$

where the exact lower and upper bounds for $P(X = x)$ are denoted by $\underline{P}(x)$ and $\overline{P}(x)$, respectively. Furthermore, $^*$ denotes the approximations of the exact lower and upper bounds with the LBBN method. $Q$ is the number of summands. In our experiments, the number of states $s$ is the same for all nodes, and, thus, $Q$ equals $2 \cdot n \cdot s$, where $n$ is the number of nodes in the network.

Table 5: Average MSE for L2U, SV2U, IPE, and LBBN Tested With Randomly Generated Multi-connected Networks.

|                 | L2U   | LBBN  | SV2U  | IPE   |
| --------------- | ----- | ----- | ----- | ----- |
| Net 2stat (1.2) | 0.009 | 0.024 | 0.048 | 0.068 |
| Net 2stat (1.4) | 0.012 | 0.031 | 0.077 | 0.189 |
| Net 2stat (1.6) | 0.011 | 0.033 | 0.086 | 0.161 |
| Net 4stat (1.2) | n/a   | 0.063 | n/a   | n/a   |
| Net 4stat (1.4) | n/a   | 0.067 | n/a   | n/a   |

In Table 5, these results can be compared to the results for L2U, IPE, and SV2U published in [Ide and Cozman, 2008]. As we can see, our simple LBBN method based on standard Bayesian networks is the second best method in terms of accuracy, and can, therefore, directly compete with L2U whose accuracy is better but whose computational costs per node are higher (see previous section). Furthermore, the last two rows of Table 5 show that our method can provide reasonable good approximations for non-binary networks, where none of the other approximation methods can provide a solution. Note that this is due to the fact that the other methods are, by design, not applicable to multi-connected networks which have a node with more than two states.

In the future, we plan to compare our method to other approximations for large networks, in particular to the GL2U [Alessandro Antonucci and de Campos, 2008] which also provides approximations for general network structures.

## 6　CONCLUSIONS

In this article, we presented a novel way to propagate lower bounds on conditional probabilities in Bayesian networks. We focused on prognostic inference and proved that our method guarantees to provide outer approximations which we showed to be competitive concerning accuracy to existing approximations, while providing the best computational complexity of all approximations. For the special case of prognostic inference in Bayesian networks with binary variables, our method provides exact solutions even *faster* than the fastest currently available method for exact inference in binary Bayesian networks, the 2U algorithm [Ide and Cozman, 2008]. In large networks, exact propagation is intractable—in our experiments we reached the limit of computing exact lower and upper bounds with a network having only 10 nodes and 16 edges—and other approximations are either computationally expensive, limited to certain network structures, or both. The LBBN method provides a feasible solution for lower bound propagation in large networks. Finally, we want to emphasize that our method can be run using any existing algorithm and implementation for inference in Bayesian networks. Therefore, it allows for a sensitivity analysis in any large Bayesian network at virtually no extra costs.

**Acknowledgment**

We would like to thank the anonymous reviewers of the article for their very constructive suggestions which helped us to improve the quality of this article. We also highly appreciate the comments of Sebastian Riedel.

## A Appendix

*Proof.* We will show that inequality (1) holds.

Let us denote the term $P(X_j|\Pi(X_j))$ as $p_{i_j}$, for $1 \leq j \leq r$. And, analogously, $P(X'_j|\Pi(X'_j))$ as $p'_{i_j}$. Finally, the term $P(X_{r+1}|\Pi(X_{r+1}))$ is written short as $b$, and $P(X'_{r+1}|\Pi(X'_{r+1}))$ is written analogously short as $b'$. We enumerate the states $S_j$ from 1 to $s_j$, and the states from $S'_j$ from 1 to $s_j + 1$, whereas $s_j + 1$ corresponds to state $N$.

For the following proof, we will exploit the following properties: For all $j \in \{1, \ldots, r\}$ it holds that $\sum_{i_j=1}^{s_j} p_{i_j} = 1 = \sum_{i_j=1}^{s_j+1} p'_{i_j}$. Furthermore, note that $b$ is dependent on $i_1, i_2, \ldots i_r$, and $p_{i_j}$ is also dependent on $i_1, i_2, \ldots i_{j-1}$. Furthermore, it holds that $\forall i_j \in \{1, \ldots, s_j\} : p_{i_j} \geq p'_{i_j} \wedge b \geq b'$. It also holds that for $i_j \neq s_j + 1$ and $1 \leq d < j$, then $p'_{i_j}(i_d = s_d + 1) \leq p'_{i_j}(i_d \neq s_d + 1)$. Moreover, $b'(i_j = s_j + 1) \leq b'(i_j \neq s_j + 1)$.

Assuming the above properties holds, we can state

$$\sum_{i_1=1}^{s_1} p_{i_1} \sum_{i_2=1}^{s_2} p_{i_2} \ldots \sum_{i_r=1}^{s_r} (p_{i_r} \cdot b) \geq$$
$$\sum_{i_1=1}^{s_1+1} p'_{i_1} \sum_{i_2=1}^{s_2+1} p'_{i_2} \ldots \sum_{i_r=1}^{s_r+1} (p'_{i_r} \cdot b').$$

Proof by induction over $r$.
Induction basis, r = 1:

$$\sum_{i_1=1}^{s_1} p_{i_1} \cdot b \geq \sum_{i_1=1}^{s_1+1} p'_{i_1} \cdot b',$$

This inequality holds, since in the term of the right hand side more mass of $p_{i_1}$ is assigned to the state of $b'$, for which $b'$ is minimal.

Induction step, from $r$ to $r + 1$:
To show is that

$$\sum_{i_1=1}^{s_1} p_{i_1} \sum_{i_2=1}^{s_2} p_{i_2} \ldots \sum_{i_{r+1}=1}^{s_{r+1}} (p_{i_{r+1}} \cdot b) \geq$$
$$\sum_{i_1=1}^{s_1+1} p'_{i_1} \sum_{i_2=1}^{s_2+1} p'_{i_2} \ldots \sum_{i_{r+1}=1}^{s_{r+1}+1} (p'_{i_{r+1}} \cdot b').$$

Let us denote

$$b^{new} := \sum_{i_{r+1}=1}^{s_{r+1}} p_{i_{r+1}} \cdot b$$

and

$$b'^{new} := \sum_{i_{r+1}=1}^{s_{r+1}+1} p'_{i_{r+1}} \cdot b'.$$

Using the same argument like in the basis, one can see that
$$b^{new} \geq b'^{new}.$$
We now have to show that $b'^{new}(i_j = s_j + 1) \leq b'^{new}(i_j \neq s_j + 1)$, for all $1 \leq j \leq r$. This holds since, $b'(i_j = s_j + 1) \leq b'(i_j \neq s_j + 1)$, and thus $b'^{new}(i_j = s_j + 1)$ is a weighted average with more weight on the smallest value of $b'$. We can now use the inductive assumption, to conclude that

$$\sum_{i_1=1}^{s_1} p_{i_1} \sum_{i_2=1}^{s_2} p_{i_2} \ldots \sum_{i_r=1}^{s_r} (p_{i_r} \cdot b^{new}) \geq$$
$$\sum_{i_1=1}^{s_1+1} p'_{i_1} \sum_{i_2=1}^{s_2+1} p'_{i_2} \ldots \sum_{i_r=1}^{s_r+1} (p'_{i_r} \cdot b'^{new}).$$

□